# Towards the Automated Segmentation of Epicardial and Mediastinal Fats

## A Multi-Manufacturer Approach Using Intersubject Registration and Random Forest

É. O. Rodrigues, A. Conci
Institute of Computing
Universidade Federal Fluminense
Niterói, Brazil
erickr@id.uff.br, aconci@ic.uff.br

F. F. C. Morais
Institute of Medicine / Pró-Laudo
Universidade Federal do Rio de Janeiro
Rio de Janeiro, Brazil
felipe.morais@pro-laudo.com.br

M. G. Pérez
Fac. de Ing. en Sist. Electr. e Ind.
Universidad Técnica de Ambato
Ambato, Ecuador
maria.espanya@gmail.com

***Abstract*— The amount of fat on the surroundings of the heart is correlated to several health risk factors such as carotid stiffness, coronary artery calcification, atrial fibrillation, atherosclerosis, cancer incidence and others. Furthermore, the cardiac fat varies unrelated to the overall fat of the subject, and, therefore, it reinforces the quantitative analysis of these adipose tissues as being essential. Clinical decision support systems are computer programs capable of evaluating information and providing a corresponding diagnosis or data to complement the physicists' analyses. The aim of this work is to propose a method capable of fully automatically segmenting two types of cardiac adipose tissues that stand apart from each other by the pericardium on CT images obtained by the standard acquisition protocol used for coronary calcium scoring. Much effort was devoted to promote minimal user intervention and ease of reproducibility. The methodology proposed in this work consists of a registration, which will roughly adjust input images to a standard, an extraction of features related to pixels and their surrounding area and a segmentation step based on data mining classification algorithms that define if an incoming pixel is of a certain type. Experimentations showed that the achieved mean accuracy for the epicardial and mediastinal fats was 98.4% with a mean true positive rate of 96.2%. In average, the Dice similarity index was equal to 96.8%.**

## I. Introduction

The cardiac epicardial and mediastinal (also termed pericardial) fats are correlated with several cardiovascular risk factors [1,2]. At the present, three techniques (i.e., modalities) appear suitable for quantification of these adipose tissues, namely Magnetic Resonance Imaging (MRI), Echocardiography and Computed Tomography (CT). Each of these modalities have been used in several medical studies in the literature [3-5]. However, computed tomography provides a more accurate evaluation of fat tissues due to its higher spatial resolution if compared to ultrasound and MRI [6]. In addition, CT is also widely used for evaluating coronary calcium score [5].

The automated quantitative analysis of epicardial and mediastinal fats may add a prognostic value to cardiac CT trials with an improvement on its cost-effectiveness. Moreover, automation can reduce the variation introduced by different observers. In fact, evaluating these data by direct user interaction is highly prone to inter and intra-observer variability. Thus, evaluated samples may not be associated to a unified common sense of segmentation. Iacobellis et al. [7] have shown that the epicardial fat thickness and coronary artery disease correlate independently of obesity, fact that supports the individual segmentation of these adipose tissues rather than merely and simply estimating that volume based on the patient overall fat.

In this work, we define the fat located within the epicardium as epicardial fat, corroborating with the majority of the published works [8-11]. Furthermore, by following the same "first outer anatomical container" logic, we conclude that mediastinal fat is the best definition for the fat located on the external surface of the heart or fibrous pericardium. In other words, the mediastinal fat is located within the mediastinal space as long as it is not epicardial (i.e., it is not located within the epicardium). Furthermore, we have used CT scans from two manufactures (Siemens and Philips), which configures this work as a multi-manufacturer approach.

## II. Literature Review

Some studies [11,12] associate the amount of epicardial adipose tissue to the progression of coronary artery calcification. Schlett et al. [12] found that epicardial fat volume is nearly twice as high in patients with high-risk coronary lesions as compared to those without coronary artery calcification. Several studies also correlate other cardiovascular risk factors and outcomes to the epicardial adipose tissue volume such as myocardial infarction [13], atrial fibrillation and ablation outcome [12], carotid stiffness [14], atherosclerosis [8,9], and many others [2,13,15-17]. Furthermore, Wei-Ta et al. have also shown that high coronary artery calcium score is associated with a high general cancer incidence [18].

Furthermore, some studies address the importance of the mediastinal fat (due to the literature inconsistency some call it pericardial fat) and its correlation with pathogenic profiles, risk factors and diseases [19-21]. Some [9,14] associate the mediastinal fat, along with the epicardial fat, to carotid stiffness and others [9,15] associate both to atherosclerosis and coronary artery calcification. Sicari et al. [3] have also shown how mediastinal fat rather than epicardial fat is a cardiometabolic risk marker.

Moreover, the 16-year study of Kresten et al. [22], that assessed a total of 384 597 patients, associates a rate of 38.4% of death to the subsequent 28 days of individuals that have had their first major coronary event. They also conclude that fatal cases is slightly less associated to female individuals. Furthermore, a study realized in Jamaica ranks cardiovascular accidents as the most common cause of sudden natural death [23], reinforcing the importance of this work.

### *A. Segmentation of the cardiac adipose tissue*

Some of the first semi-automated segmentation methods for the epicardial fat were proposed since 2005. Dey et al. [24], for instance, apply a preprocessing step to remove all other structures apart from the heart by using a region growing strategy. Therefrom, an experienced user is required to scroll through the slices to place from 5 to 7 control points along the pericardium border if the it is visible. Catmull-Rom cubic spline functions are automatically generated to obtain a smooth closed pericardial contour. Finally, since the epicardial fat is inside this contour it is simply accounted by thresholding. In [25] a method for the segmentation of abdominal adipose tissue is proposed. The work of Kakadiaris et al. [26] have further extended the method introduced by Pednekar et al. [25] to the segmentation of the epicardial fat.

Coppini et al. [6] focused on reducing the user intervention. On their method, an expert is still necessary to scroll through the slices between the atrioventricular sulcus and the apex in order to place some control points on the pericardium. The amount of essential points is not described clearly. Nevertheless, the amount of slices to be analyzed is apparently lesser than the ones on the method proposed by Dey et al. [24]. They also present their solution in a 3D space and claim that Dey et al [24] do not. The overall focus of their work was to describe their method mathematically. However, the work lacks heavily on describing the general accuracy of their method.

Barbosa et al. [27] proposed a more automated segmentation method for the epicardial fat. They use the same preprocessing method from Dey et al. [24] and further apply a high level step for identification of the pericardium by tracing lines originating from the heart's centroid to the pericardium layer and interpolating them with a spline. Although such approach may be interesting, of simple complexity and highly applicable for virtually any proposed method in the field, the reported results are not impressive. Only 4 out of 40 images were correctly segmented in a fully automatic way.

Shahzad et al. [28] proposed, to the extent of our knowledge, the first fully automated method for epicardial fat segmentation in 2013. Their method uses a multi-atlas based approach to segment the pericardium. The multi-atlas approach is based on registering several atlases (8 in this case) to a target patient and on fusing these transformations to obtain the final result. They selected 98 patients for testing and reported a Dice similarity coefficient of 89.15% to the ground truth and a low rate of approximately 3% of unsuccessful segmentations. Nevertheless, they did not provide any measurements of the overall processing time.

Ding et al. [29] proposed in 2014 an approach that is similar to the method of Shahzad et al. [28]. They segment the pericardium based on an atlas, which consists of a minimization of errors after applying transformations to the atlas along with an active contour approach. Their mean Dice similarity coefficient was 93% and they claim that their result was achieved in 60 seconds on a simple personal computer. Although their segmentation seems to be the most precise in the literature, the reported computing time is poorly described. We consider 60 seconds too fast for segmenting and transforming an entire scan, which consists of roughly 50 images. They also present a work [30] that segmented the aorta instead of the pericardium and compare their achieved time (60 seconds) to the 15 minutes of the former. If these 60 seconds correspond to just the time it takes for the algorithm to minimize the transformations, then this comparison is not feasible. Furthermore, they report that on their approach the atlases' images were pre-aligned to a standard orientation and, thus, there is a comparison with only one of the atlases to speed up the process. The remaining pericardium contour will follow the pre-aligned pattern, which is a reported limitation. Besides, they did not describe how each one of these atlases is chosen as the correct one for each possible case.

Rikxoort et al. [31] have proposed the use of the k-nearest neighbor algorithm for segmentation of the liver on CT images. The core of their method consists of a voxel labeling procedure: (1) for every voxel in the test set a number of numerical values (a features vector) is computed and (2) a statistical classifier, trained on previously extracted features vectors, evaluates if the analyzed voxel is or is not part of the liver. The approach proposed on this work is similar to [31]. However, we have extracted more features, evaluated various classifiers instead of one, applied it to a different problem and combined the classification approach to an intersubject registration.

## III. Proposed Methodology

The proposed automatic segmentation is based on two main principles, namely (1) a registration and (2) a classification step. For the whole segmentation process we have used CT images on its respective range of fat (from -200 to -30 HU [28,32-34]). The reason for choosing such interval was for directly quantifying the amount of fat an individual has by counting the pixels of the image. However, we believe that our methodology can be successfully applied to other ranges and also other modalities.

Image registration can be defined as the process of matching characteristics from images in order to search for alignments that minimize the variation between overlapping pixels or areas of pixels [35]. Such processes are included on panoramas assemblages, medical images, time series alignments [36,37] and on many others tasks. Registration is also alternatively treated as an optimization problem with the goal of finding the spatial mapping that will bring images, parts of them, or even a combination of these parts into minimal variation.

Machine learning algorithms are often divided in two main categories: (1) the supervised and (2) the unsupervised methods. The algorithm is categorized as supervised when it explicitly evaluates the class or label attribute of a training set as the predictive label desired to attach to an incoming unlabeled instance. Furthermore, when this assumption is formalized, the class attribute heavily induces the generated predictive model. However, when not formalized, the algorithm is defined as unsupervised and the class plays no heavy influence but of a normal attribute, when it is not disregarded from the training phase. Classification algorithms are always categorized as supervised learning methods while clustering algorithms are often unsupervised.

### *A. Image Registration*

The intent of this work is to autonomously perform the registration of several patients despite of the scaling and actual positioning of their hearts on their CT images. Therefore, there is no likelihood for manual placement of any common landmark such as several works have proposed [24-26,38]. Thus, the remaining alternative is to autonomously find a selected landmark. Hence, in other words, the parameters of our transformation are searched for and determined by finding an optimum of some function on the search space.

The subject type of our solution cannot be intrasubject, since we do not want to align structures of a patient based just on its own information. As a matter of fact, we want to align the structures of several distinct patients to some extent. Our approach is based on affine transformations since the required registration to be applied have to support scaling and translations. Summarily, according to the definitions of Maintz et al. [36], our proposed registration approach is categorized as of intrinsic nature where the parameters are search for and the transformation applied is affine. The main steps of our registration are shown on Figure 1.

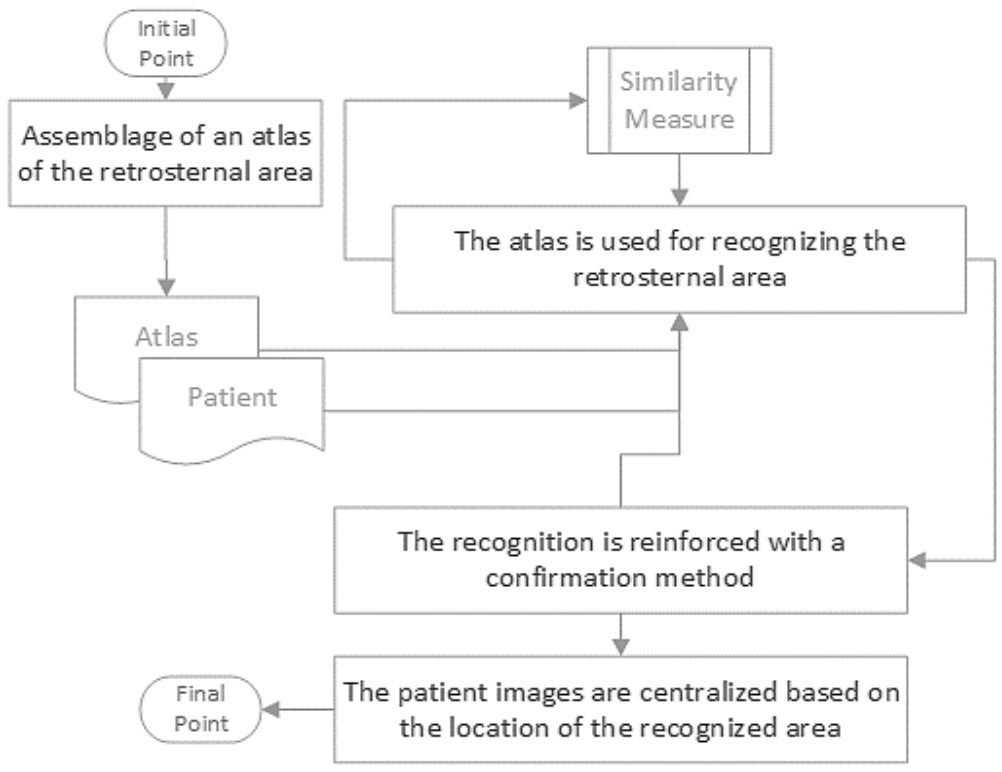

Fig. 1. Main steps of the proposed registration.

At first, an atlas was assembled by binary thresholding (between pixel values equal to and higher than 0) 10 retrosternal areas of 10 randomly chosen and manually aligned distinct patients and combining these images with an arithmetic mean (on the fat range only) as shown on Figure 2. We have selected just one slice from each patient and they were as next as possible to the shoulders. It is important to highlight that the retrosternal area, which was the chosen common landmark, is the region located on the back of the sternum. Taking the Figure 4 as reference, the central point of the retrosternal area is illustrated by the green plus sign whereas the atlas image is shifted to the blue color.

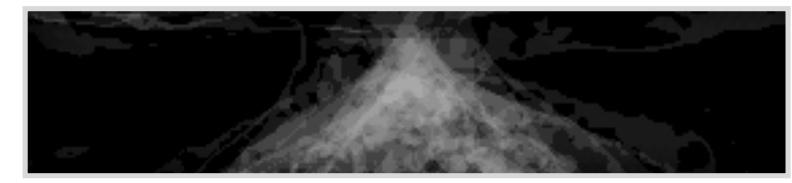

Fig. 2. Atlas of the retrosternal area.

Moreover, as soon as the atlas is assembled, the next step consists of using the same to recognize the retrosternal area of each patient. For doing so, a similarity measure can be used. Therefrom, the recognition approach consists of moving the atlas through one single slice of each patient and availing each position with a similarity score. The weighted mutual information (WMI) was empirically selected (with a successful rate of recognitions of 70%) for computing the similarity [39]. The WMI formula is shown on the Equation (1). Furthermore, we have also tested the normalized correlation, sum of the differences and the common mutual information as similarity measures but all of them provided worse results.

$$WMI_{y,x}(F,M,g) = \left(\sum_{f\in M\rightarrow F}\sum_{m\in M}\frac{1}{|f-m|+1}\rho_{FM}(f,m)\log_g\frac{\rho_{FM}(f,m)}{\rho_F(f)\rho_M(m)}\right) \quad (1)$$

The successful rate of 70% achieved by the WMI is relatively high but did not sufficiently high as we expected. We defined as successful every recognition of the retrosternal area that was visually correct, regarding a slightly variation on the positioning. Therefore, to enhance this rate we have combined the atlas approach with a heuristical confirmation method. The confirmation method was thought to reinforce the position chosen by the atlas scoring. Its heuristic is simple and straightforward. Given a small rectangle area of pixels $A$ at the center of the recognized retrosternal area, there should be two points $p_l$ and $p_r$ that belongs to $A$ and that continuously move only through fat pixels (the ones that are not background, or black) on the left-bottom and right-bottom direction respectively until they hit convergence. The first image on Figure 3 stands for a retrosternal recognition done by the atlas scoring approach, the center of the retrosternal area is pointed by the pink arrow and the rectangular area $A$ is represented in yellow. The second instance on Figure 3 depicts the reason why, due to the discontinuity of fat depots in some cases, an area $A$ should be considered to evaluate the starting points. Both images are displayed on the fat range only, where, as previously described, the processing of our approach took place. All the pixels that are not within this range are considered background and were colored as black (0). The non-smoothness of these images is a direct consequence of this type of displaying.

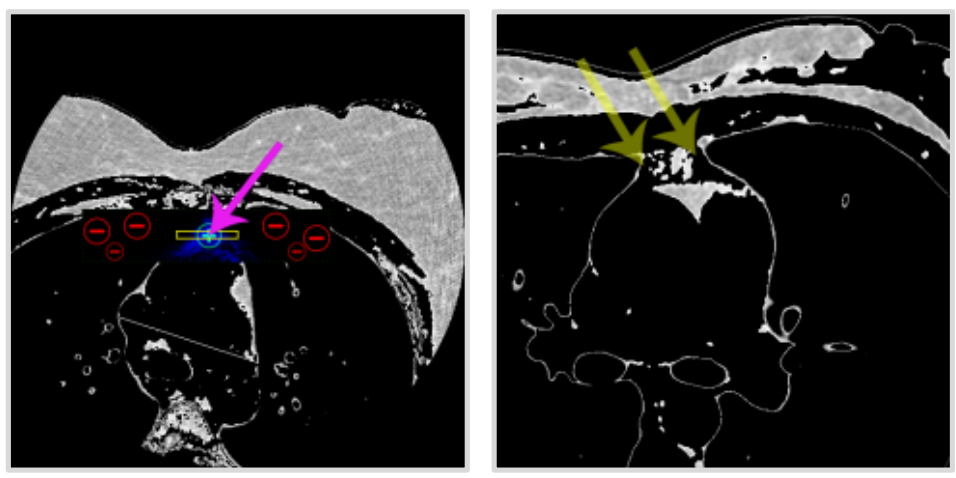

Fig. 3. Central point of the retrosternal area and descuntinuity of fat depots.

The thin slanted white lines crossing the middle of the heart on the instances of Figure 4 illustrate the binding of the two points $p_l$ and $p_r$ after their convergence. The action of moving these points on the image is just a step of the confirmation method. For better explanation, we define $w$ as being the width of the image (512 pixels in our case). The logical step that will reinforce if the selected position for the atlas is correct is defined as: (1) the white line must be within a certain width, i.e., be bigger than $0.2w$ and smaller than $0.55w$. Besides, (2) the displacement of the points must also be within a certain length, i.e., both displacements must have at least half-length of the other. Finally, (3) both points must also be within a certain distance from the starting point, i.e., they must be away from the starting point by at least $0.2w$ pixels. If this confirmation fails, the settlement of the atlas (i.e., the recognition of the retrosternal area) and the confirmation method must be redone jointly for every evaluated position. It is important to highlight that depending on the interpolation used after the rescaling process (according to its DICOM file), some gaps may appear on the image, mainly on the surrounding line of fat of the heart. These produced gaps should be properly accounted by the confirmation method by, for instance, jumping a relative number of non-fat pixels when moving $p_l$ and $p_r$. In other words, if the interpolation usually lefts a gap of 2 pixels in the images, then the points should be forced to move 1 pixel and, if not possible, they should try to move 2.

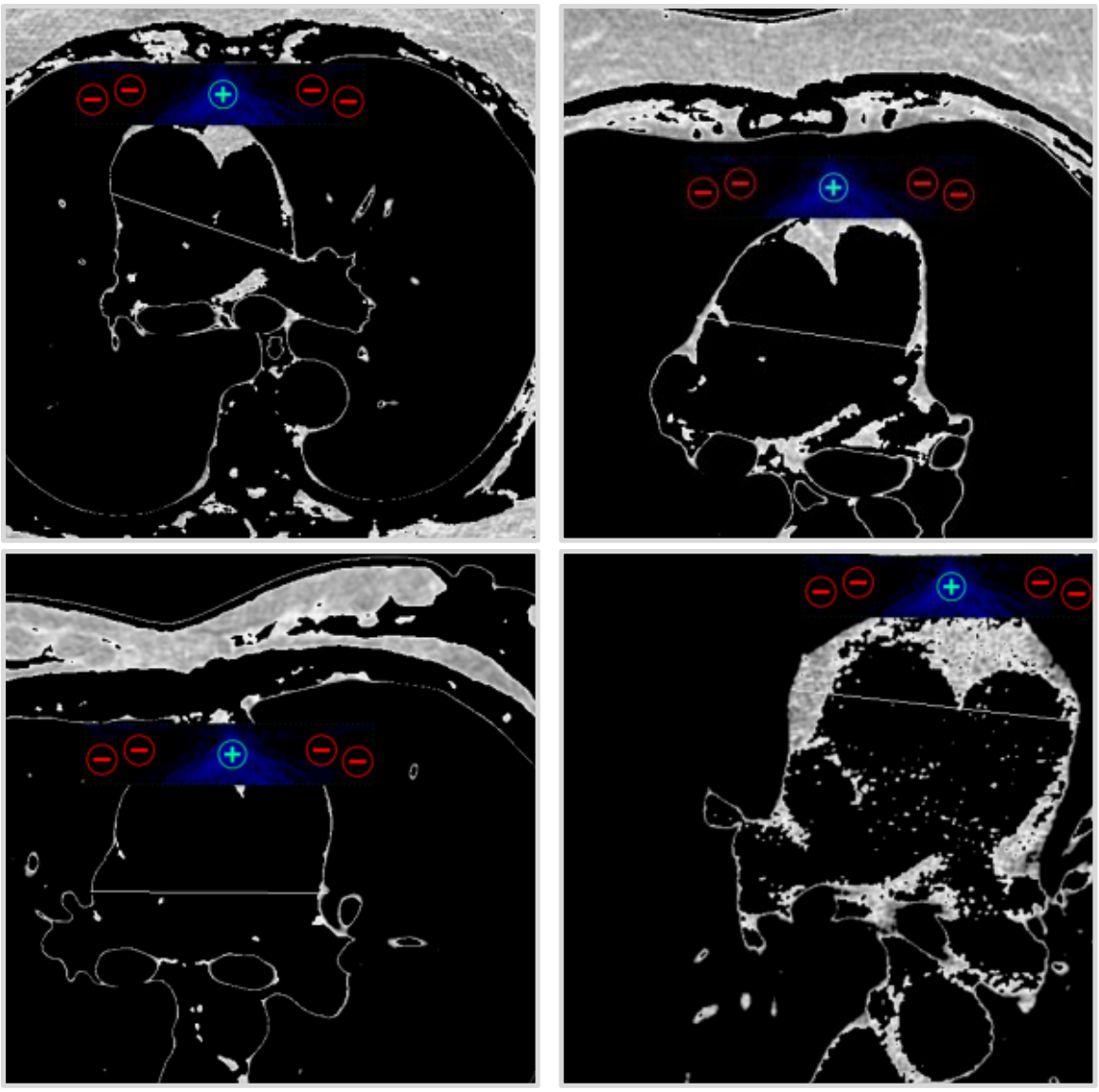

Fig. 4. Binding of the two points $p_l$ and $p_r$ and recognition of the retrosternal area.

By combining the WMI and the confirmation method, the successful rate raised from 70% to 100%. That is, among all the 82 instances we have assessed, all of them had their retrosternal area properly recognized. In our case, prior to the recognition step, the images were rescaled according to the pixel spacing data on the DICOM file. Therefore, the recognition was done on images that were already rescaled. Subsequently to the retrosternal area recognition and based on that information, all the images of the patient are translated to a common centralized point, which standardizes the positioning of the heart. Although we have rescaled the images prior to the recognition, we confirmed that the proposed registration works for varying scales as well. Furthermore, the recognition step just need to be done once for each patient, the same transformation applied to a single slice can be applied to the remaining. The 3 instances shown on Figure 5 represent 3 distinct patients before (first row) and after (second row) the registration. The images after the registration are represented on the fat range while the others are on the (-200,500) HU range.

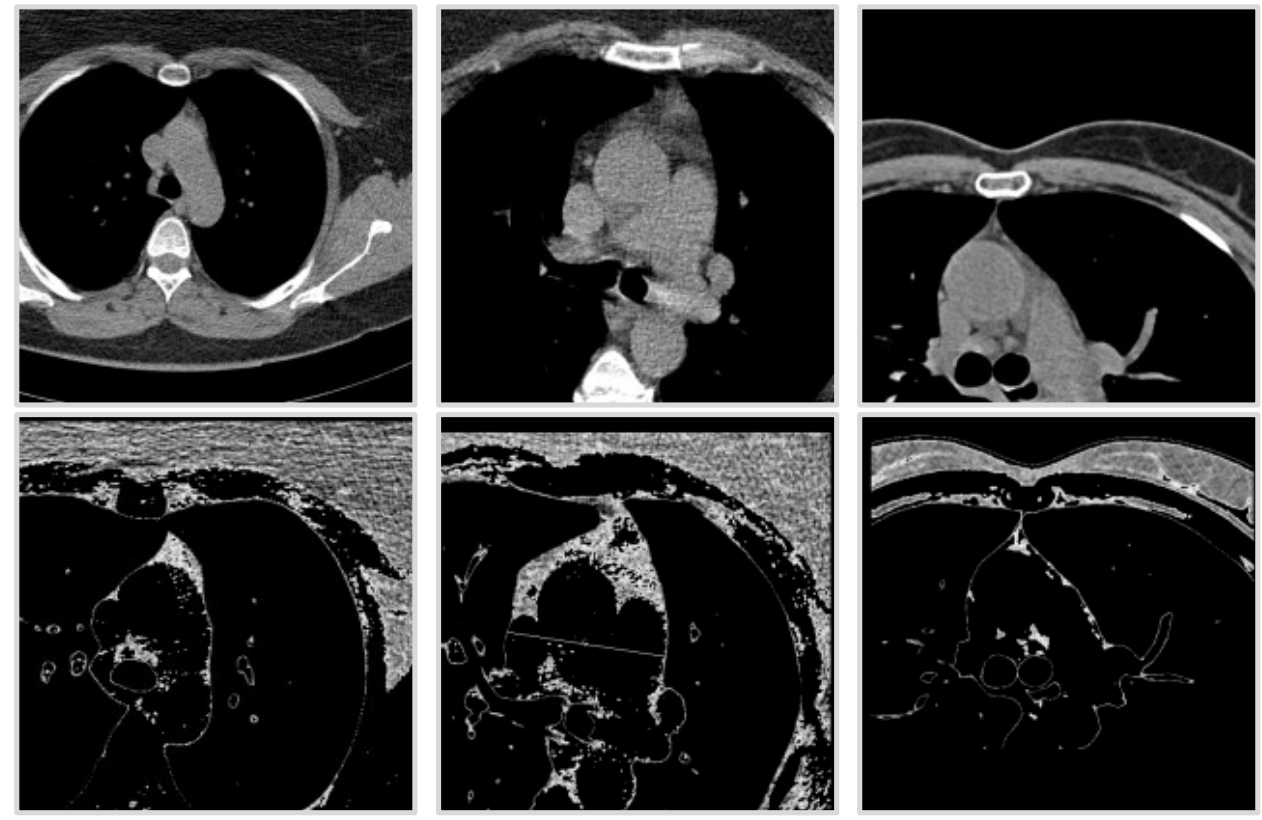

Fig. 5. Results of the proposed registration.

## *B. Segmentation*

We define the act of segmenting an image using classification algorithms as classified segmentation. This procedure is also termed in the literature as pixel classification or probability based segmentation [40]. Several other approaches for image segmentation employ commonly used procedures such as a simple thresholding, clustering, edge detection, level set, active contour, region growing, atlas matching and many others [28,30,40-42].

The classifying segmentation can be viewed as a simple iteration through a set of pixels or voxels of an image or 3D model where a set of features (or characteristics) related to the iterated pixel, voxel or surrounding area is extracted. These extracted features are illustrated as the variable **f** on Figure 6. The set of features is usually called features vector. Several vectors of this type will usually compose a dataset that is provided as input to a classification algorithm.

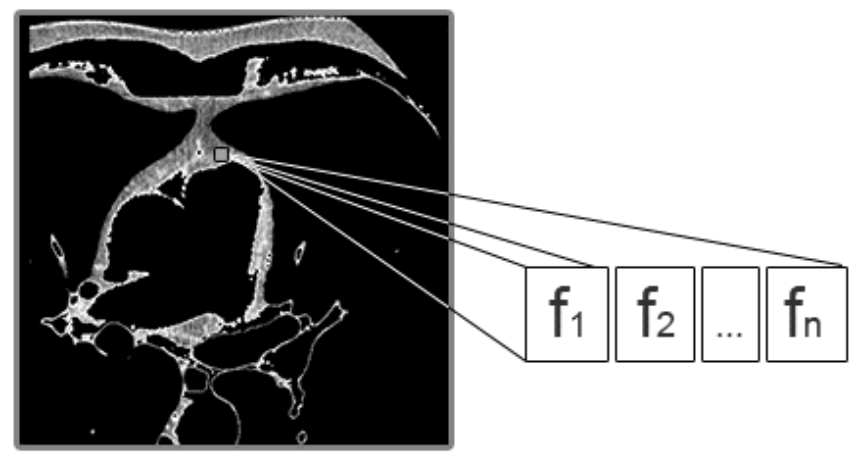

Fig. 6. Features being extracted from a pixel.

In order to generate a concise predictive model we need to supply reliable data for the training phase of the classification algorithm, which is strictly necessary for manually segmenting an incoming patient. Therefore, two specialists, one being a physicist and the other being a computer scientist, have manually segmented the epicardial and mediastinal adipose tissues of 20 patients or CT scans (10 male and 10 female). Thus, our ground truth contains approximately 1000 manually segmented cardiac CT images. It is important to highlight that previously to the manual segmentation the images were already registered by our registration method. Our ground truth is available at [43]. The black grey value (0) still represents the background and these pixels are excluded from the feature extraction.

The generated ground truth is used as source for the selected features to be extracted. Posteriorly, some classification algorithms should be selected to train on this extracted data. Therefore, our whole classification approach consists of the three main steps: (1) extracting the features, (2) training the predictive model and (3) classifying an incoming CT scan. The steps (1) and (2) do not need to be redone every time a new incoming scan needs to be classified. In fact, if that were true, the method would take so long to converge that it would be unpractical. It do not take too much to perceive that the generated dataset is very big and that the step (1) is a very slow process. Thus, the step (3) is independent of (1) and (2) once they have already been done. These three steps are illustrated as the "Initial Points" on the following Figure 7.

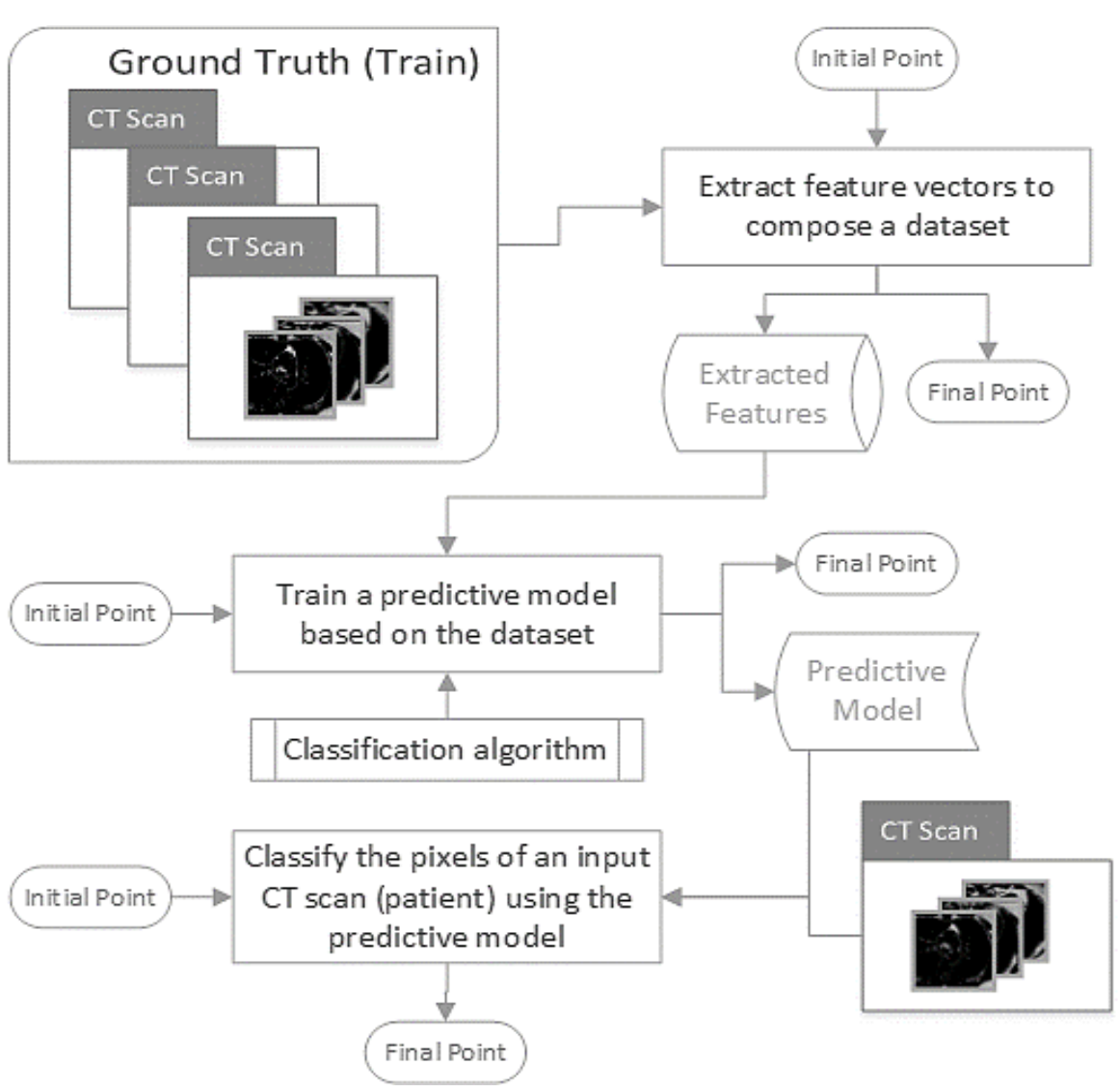

Fig. 7. The overall approach for the classifying segmentation.

We have selected as primary features: the pixel grey level and the position x, y and z, where z is the index of the slice. Besides, we have also selected the x and y positions relative to the center of gravity of the image and texture-based features from a vicinity of variable size that encapsulates the iterated pixel at its center, i.e., a surrounding window of pixel values. Some features [44] were computed from this vicinity, such as: (1) a simple arithmetic mean of the grey levels, (2) moments of the co-occurrence matrix, (3) geometrical moments of the grey values, (4) run percentage, (5) grey level non-uniformity and (6) a coefficient of smooth variation (CSV). The CSV acts as a convolution where the weights of the Kernel are based on the sup metric and on a unidimensional Gaussian filter. More details about this coefficient can be found on [45].

The reason for extracting texture-based features was due to conceiving the hypothesis that the epicardial and mediastinal fat yield a slightly difference on their texture that can be partially accounted by these features. An evidence to corroborate this hypothesis would be a well positioning of these features on a reliable ranked evaluation of attributes. In the case of decision tree algorithms, the outputted predictive tree can also be used to estimate how important and decisive is a feature on the predictive model.

## IV. Results

For the classification tasks we have used the Weka library [46]. Weka is an open-source collection of machine learning algorithms maintained by the University of Waikato. The Weka usage is twofold, it has its own graphical interface that can be used on several types of data analyses and the library can be directly imported and used on Java code. The full set of classification algorithms present in Weka up to its version 3.6.11 was selected for a speed evaluation. Some of these algorithms are, namely, the SVM, SMO, Naïve Bayes, RBFNetwork, Random Trees, CJ45, J48, SPegasos, REPTree, IBk, kNN, Multilayer Perceptron and others. Among all the tested algorithms, only the (ordered by the achieved accuracy): J48Graft, Random Forest, REPTree, J48, SimpleCart, SMO, RandomTree, RBFNetwork, SPegasos, DecisionStump and NaiveBayes converged within 200 seconds. Although the J48Graft achieved a greater accuracy than the Random Forest, the latter provided a more sparse and acceptable segmentation. Preliminary results of two slices from two distinct patients (on each row) using the Random Forest algorithm are shown on Figure 8. The red color represents the epicardial while the green color represents the mediastinal fat.

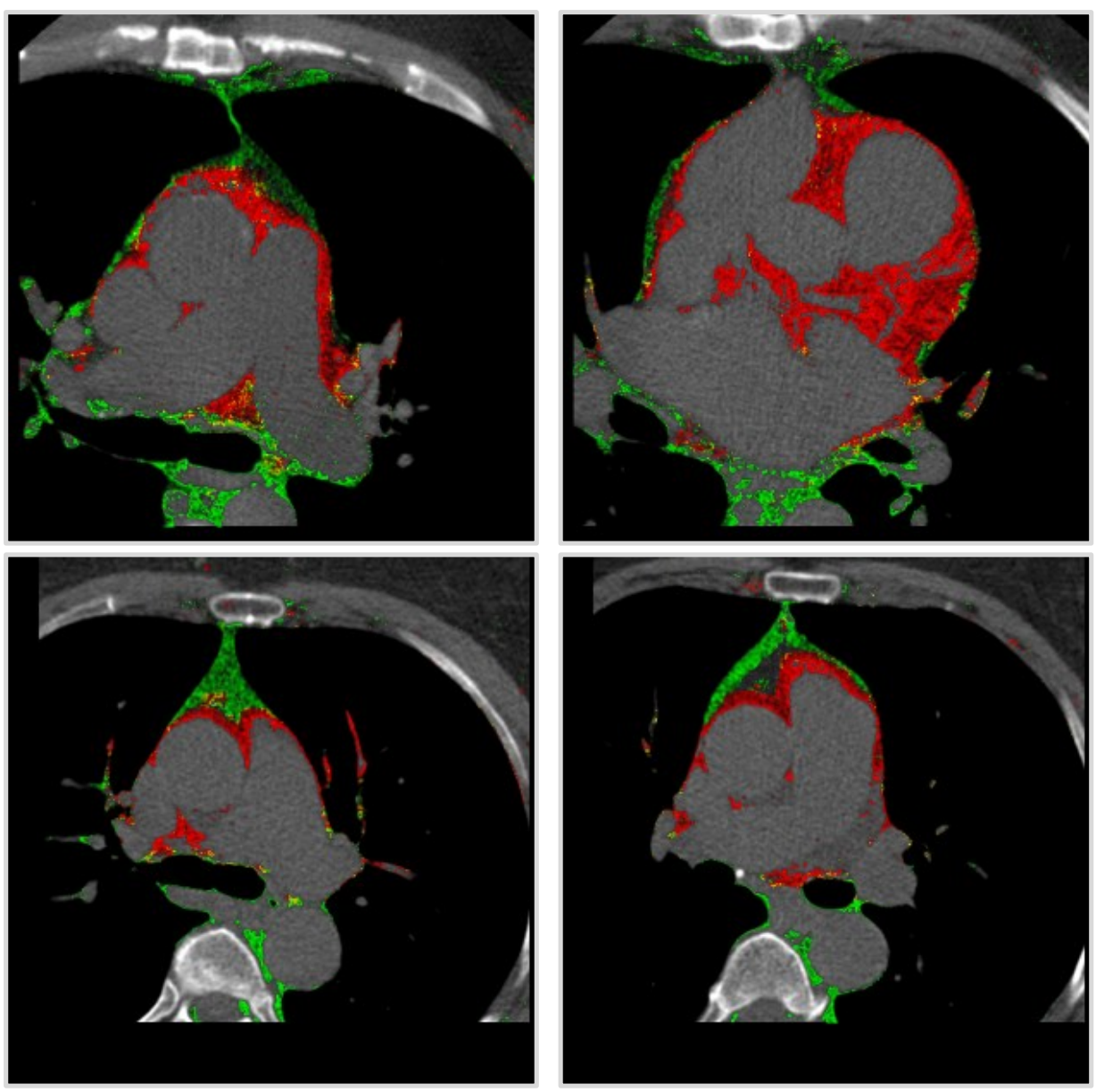

Fig. 8. Automatic segmentation of the cardiac fats on [-200,500] HU.

The efficiency of a predictive model is usually evaluated by a test mode. The two most commonly applied techniques for evaluating the accuracy of a predictive model are the split and the cross validation. The split method usually takes (randomly or not) a percentage of the dataset to train the predictive model and the remaining for testing its accuracy. A $k$%-split test mode, for instance, divides $k$% of the dataset for training and use the remaining for testing. The cross validation can be viewed as a more sophisticated split method. It divides the dataset in $k$ subsamples. A single subsample is retained for evaluation of the predictive model while the remaining $k-1$ are used for training. The difference is that, on the cross validation, this process is done $k$ times while varying the retained subsample and the final statistics are the average of the $k$ tests.

A database of approximately 3 gigabytes originated from 20 patients (ground truth) was generated for extensively evaluating our proposed segmentation approach using the Random Forest algorithm. This database is also directly available on the Weka's arff format at [43]. For this occasion, we considered the 66% split evaluation and the 10-fold cross validation test modes. The difference between the two is not expressive due to the huge amount of instances on the dataset. Table 1 contains the accuracies and the confusion matrixes of the Random Forest with standard parameters (-I 10 -K 0 -S 1) over 20 patients, using a 25x25 vicinity size and obtained through the random 66%-split test mode, whereas the Table 2 contains the values obtained through the 10-fold cross validation.

TABLE I. RESULTS WITH 66%-SPLIT TEST MODE

| **Tissue** | **Rates** | | | | |
|---|---|---|---|---|---|
| | ***Accuracy*** | ***TP Rate*** | ***TN Rated*** | ***FP Rate*** | ***FN Rate*** |
| Epicardial | 98.3% | 98.1% | 98.4% | 1.6% | 1.5% |
| Mediastinal | 98.0% | 92.9% | 98.8% | 1.1% | 1.1% |

TP = true positive, TN = true negative, FP = false positive and FN = false negative

TABLE II. RESULTS WITH 10-FOLD CROSS VALIDATION TEST MODE

| **Tissue** | **Rates** | | | | |
|---|---|---|---|---|---|
| | ***Accuracy*** | ***TP Rate*** | ***TN Rated*** | ***FP Rate*** | ***FN Rate*** |
| Epicardial | 98.5% | 98.3% | 98.5% | 1.4% | 1.4% |
| Mediastinal | 98.4% | 94.2% | 99.1% | 0.9% | 0.9% |

TP = true positive, TN = true negative, FP = false positive and FN = false negative

The Table 3 compares the results of the four main related works. When is the case that some values are not provided by the authors, the respective cell was left blank. Furthermore, the majority of these indexes are highly subjective. None of the works provides a publicly available ground truth. Thus, we are the first to provide a publicly available ground truth for further comparison [43]. The works of Barbosa et al. [27] and Kakadiaris et al. [26] are semi-automated, while the works of Shahzad et al. [28] and Ding et al. [29] are fully automatic. The first column indicates the rate of successful automatic segmentations (usually observed). All these four works proposed methods for segmenting just the epicardial fat and, therefore, we compare just our epicardial fat segmentation on this table.

TABLE III. COMPARING THE EPICARDIAL SEGMENTATION

| **Authors** | **Evaluated Indexes** | | |
|---|---|---|---|
| | ***Successful A.S.[a]*** | ***Dice*** | ***TP Rate*** |
| Barbosa et al. | 10% (4/40) | - | - |
| Kakadiaris et al. | - | - | 85.6% |
| Shahzad et al. | 96.9% (95/98) | 89.15% | - |
| Ding et al. | - | 93.0% | - |
| This work (epicardial) | 100% (82/82) | 97.9% | 98.3% |

[a] A.S. = automatic segmentations

## V. CONCLUSION

The appliance of classification algorithms on image segmentation is highly prone to success and may surpass many usual segmentation methods on several aspects. Random Forest is one of the most well rated decision-tree algorithms and was the most efficient in our analysis. We have also concluded that decision tree algorithms provided much better performance over neural networks and function-based classification algorithms. The achieved mean accuracy for the epicardial and mediastinal fats was 98.4% with a mean true positive rate of 96.2%. In average, the Dice similarity index was 96.8%.

The current weakness of our approach is the processing time. Currently, with a huge set of extracted features and an optimized heuristical segmentation, the algorithm still takes some hours to fully segment a patient on a simple personal computer with a dual core processor and 4 gigabytes of memory. What could diminish this huge processing time would be an extensive evaluation of the features in order to select just the most important and disregard the remaining. That kind of selection would speed up the feature extraction, reduce the outputted dataset which, consequently, speeds up the classification step. Algorithms for attribute evaluation consume a significant time until reaching convergence, especially on a large dataset. Due to that fact, we were still not able to extensively evaluate such matters.

## ACKNOWLEDGMENT

E. O. Rodrigues wants to thank CAPES. A. Conci wants to thank the CNPq project 302298/2012-6.